\documentclass[11pt,letterpaper]{article}

\usepackage[preprint]{acl}

\usepackage{times}
\usepackage{latexsym}
\usepackage[T1]{fontenc}
\usepackage[utf8]{inputenc}
\usepackage{microtype}
\usepackage{graphicx}
\usepackage{amsmath}
\usepackage{amssymb}
\usepackage{booktabs}
\usepackage{float}
\usepackage{dsfont}

\title{Calibrated Selective Fact-Checking via Evidence Chain Evaluation}
\author{Dekun Yang \\
  Zhejiang University \\
  \texttt{pauliyangwork@gmail.com}}

\begin{document}
\maketitle

\begin{abstract}
Large language models (LLMs) can achieve strong fact-checking accuracy, yet forced binary decisions conceal a critical reliability problem: systems may issue confident verdicts even when supporting evidence is weak, sparse, or internally inconsistent. We address this issue through \textbf{Evidence Chain Evaluation (ECE)}, a selective fact-checking framework that permits abstention via an \textit{uncertain} verdict instead of requiring a true/false decision for every claim. The evaluated system is a tool-using verification agent that gathers evidence through web search, scholarly search, and executable checks, and then returns a structured verdict with confidence and source-level metadata. On ECE-Bench, ECE achieves 91.6\% standard accuracy, 93.7\% coverage, and 97.8\% selective accuracy on answered claims. Although ECE does not outperform the strongest retrieval baseline on aggregate calibration metrics such as Expected Calibration Error, Brier score, or AURC, it delivers a clear selective-prediction trade-off: the system maintains very high accuracy on answered claims while deferring 6 of 95 cases. These deferred cases are concentrated in lower-reliability evidence settings (5/6 at source level L4), supporting the view that abstention functions as a safety-oriented mechanism for handling epistemically weak evidence. Code is available at \url{https://github.com/cheshireyang/ECE.git}.
\end{abstract}

\section{Introduction}

Large language models (LLMs) have become increasingly capable fact-checking assistants, especially when paired with retrieval or tool use \cite{guo2022survey, thorne2018fever}. However, strong top-line accuracy alone is insufficient for trustworthy deployment. In realistic verification settings, a system should not only answer correctly when evidence is strong, but also refrain from overconfident judgments when evidence quality is poor, sparse, or contradictory \cite{xiong2024can, kadavath2022language}.

Selective prediction provides a natural framing for this requirement \cite{geifman2017selective, el2010foundations}. Rather than forcing a prediction for every claim, the system may abstain when it lacks adequate support. In high-stakes verification pipelines, this behavior can be preferable to an incorrect but confident answer.

This paper studies \textbf{Evidence Chain Evaluation (ECE)} as a selective fact-checking system. Importantly, the evaluated system is not a dense-retrieval or graph-construction pipeline. Instead, it is a tool-routed verification agent that gathers evidence through web search, scholarly search, and executable checks, then returns a structured verdict with a confidence score and a source-level tag. Clarifying this point is essential: the paper's contribution should be grounded in the implemented system rather than in an idealized architecture.

Our main empirical finding is therefore more precise than a broad calibration claim: ECE achieves \textbf{97.8\% selective accuracy at 93.7\% coverage} on ECE-Bench. This result is best understood as a selective-prediction trade-off. Relative to forced-answer baselines, ECE yields coverage on a small number of difficult claims in order to preserve high answered-claim accuracy. At the same time, the strongest retrieval baseline remains superior on standard accuracy and on aggregate calibration metrics.

The contributions of this paper are:
\begin{itemize}
    \item A faithful description of the implemented ECE system as a tool-using selective fact-checking agent.
    \item A benchmark evaluation framed around coverage, selective accuracy, and risk--coverage behavior.
    \item An analysis showing that ECE's abstentions are concentrated mainly in low-reliability source settings, supporting a safety-oriented deferral interpretation.
\end{itemize}

\section{Related Work}

\subsection{Fact-Checking and Verification}

Automated fact-checking research spans benchmark construction \cite{vlachos2014fact, thorne2018fever, aly2021feverous}, end-to-end verification models, and broader surveys of misinformation detection \cite{guo2022survey}. Prior work consistently highlights that strong verdict accuracy does not eliminate concerns about hallucination, evidence quality, and source credibility \cite{huang2023survey, maynez2020faithfulness, chen2023combining, pennycook2021psychology}.

\subsection{LLM Confidence and Calibration}

Recent work shows that LLM confidence is only imperfectly aligned with correctness \cite{kadavath2022language, xiong2024can}. Proposed remedies include confidence prompting \cite{lin2022teaching}, linguistic calibration \cite{band2024linguistic}, and post-hoc analysis \cite{guo2017calibration, jiang2021can}. Together, these studies motivate evaluating not only accuracy but also whether confidence behaves sensibly under uncertainty.

\subsection{Selective Prediction}

Selective prediction studies the trade-off between answering more examples and making fewer mistakes on the examples that are answered \cite{geifman2017selective, geifman2019selectivenet, el2010foundations}. The same perspective has been applied to NLP settings such as ambiguous question answering \cite{kamath2020selective}. This paper adopts that framing for fact-checking with abstention.

\subsection{Retrieval-Augmented Fact-Checking}

Retrieval augmentation improves factual grounding by exposing the model to external evidence \cite{lewis2020retrieval, gao2024retrieval}. Tool-use systems such as WebGPT and Toolformer further show that models can improve reliability by actively interacting with search or other external tools \cite{nakano2021webgpt, schick2023toolformer}. Our evaluated ECE system belongs to this general family of tool-using verification agents, while focusing specifically on selective fact-checking behavior.

\section{Background: Selective Fact-Checking Metrics}

Let $\mathcal{D} = \{(c_i, y_i)\}_{i=1}^{n}$ be a benchmark of claims, where each claim $c_i$ has gold label $y_i \in \{\mathrm{true}, \mathrm{false}\}$. For each claim, the fact-checking system outputs a verdict $\hat{v}_i \in \{\mathrm{confirmed}, \mathrm{refuted}, \mathrm{uncertain}\}$ and a confidence score $p_i \in [0,1]$.

To mathematically formalize correctness, we define a matching function $M(\hat{v}_i, y_i) \in \{0, 1\}$ that maps a system's verdict against the gold label:
\begin{equation}
M(\hat{v}_i, y_i) = 
\begin{cases} 
1 & \text{if } (y_i = \mathrm{true} \land \hat{v}_i = \mathrm{confirmed}) \\
  & \text{or } (y_i = \mathrm{false} \land \hat{v}_i = \mathrm{refuted}) \\
0 & \text{otherwise}
\end{cases}
\end{equation}
Notice that an $\mathrm{uncertain}$ verdict strictly evaluates to $0$ (incorrect) under this definition, as it fails to commit to the true label.

\subsection{Standard Accuracy, Selective Accuracy, and Coverage}

In our evaluation script, standard accuracy treats an \textit{uncertain} verdict as a non-correct outcome. Formally,
\begin{equation}
\operatorname{Acc}_{\mathrm{std}} = \frac{1}{n} \sum_{i=1}^{n} M(\hat{v}_i, y_i).
\end{equation}

Let $\mathcal{A} = \{i : \hat{v}_i \neq \mathrm{uncertain}\}$ denote the answered subset (i.e., claims where the system committed to a binary verdict). Selective accuracy and coverage are then defined as:
\begin{align}
\operatorname{Acc}_{\mathrm{sel}} &= \frac{1}{|\mathcal{A}|} \sum_{i \in \mathcal{A}} M(\hat{v}_i, y_i), \\
\operatorname{Coverage} &= \frac{|\mathcal{A}|}{n}.
\end{align}

These metrics separate ``how often the system answers'' from ``how accurate it is when it does answer.''

\subsection{Calibration and Risk Metrics}

We report three uncertainty-related metrics from the evaluation script. First, Expected Calibration Error (ECE) measures the gap between confidence and empirical accuracy across confidence bins \cite{guo2017calibration}. Here, ECE is computed over all $n$ claims using the model's scalar confidence score $p_i$ against the binary correctness indicator $M(\hat{v}_i, y_i)$. Importantly, uncertain verdicts are retained in this calculation as predictions with their associated confidence values:
\begin{equation}
\operatorname{ECE} = \sum_{b=1}^{B} \frac{|B_b|}{n} \left| \operatorname{acc}(B_b) - \operatorname{conf}(B_b) \right|,
\end{equation}
where $B_b$ represents the set of indices for claims whose confidence falls into the $b$-th bin, $\operatorname{acc}(B_b)$ is the mean of $M(\hat{v}_i, y_i)$ in that bin, and $\operatorname{conf}(B_b)$ is the mean confidence $p_i$.

Second, we report the Brier score, a proper scoring rule over probabilistic predictions. To compute the Brier score, we map the model's output to an implied probability of the claim being true: $P(\mathrm{true}) = p_i$ if $\hat{v}_i = \mathrm{confirmed}$, $P(\mathrm{true}) = 1 - p_i$ if $\hat{v}_i = \mathrm{refuted}$, and $P(\mathrm{true}) = 0.5$ if $\hat{v}_i = \mathrm{uncertain}$. The score is the mean squared error against the binary gold label.

Third, we report area under the risk--coverage curve (AURC), computed by ordering all $n$ predictions by their confidence $p_i$ in descending order. The empirical risk is tracked as coverage expands, and AURC is the integral of this curve. Lower values indicate lower accumulated risk at high-confidence operating points. For all three metrics (ECE, Brier, AURC), lower is better.

\section{Method: Evidence Chain Evaluation as Implemented}

\subsection{Evaluated Pipeline}

\begin{figure*}[t]
\centering
\includegraphics[width=0.98\textwidth]{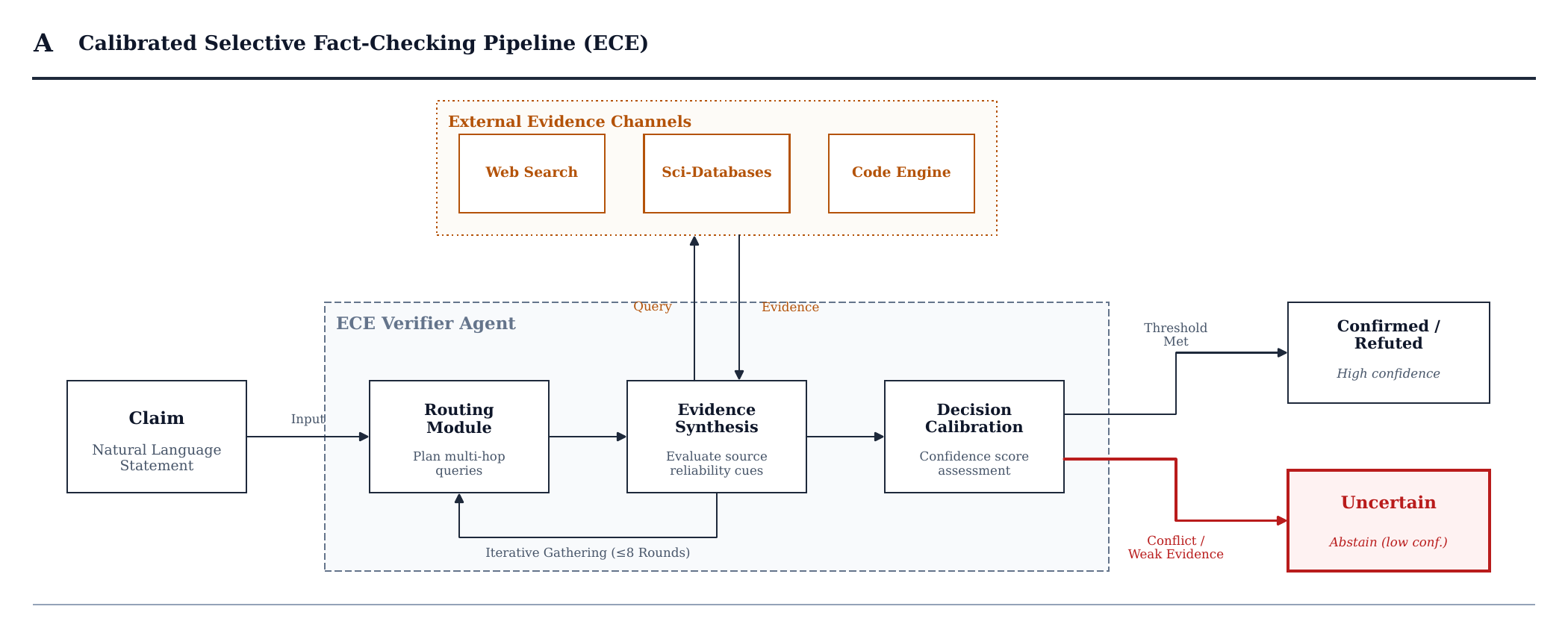}
\caption{System overview of the implemented ECE pipeline used in evaluation. A claim is processed by a tool-routed verification agent that gathers evidence through web search, scholarly search, and executable checks, then emits a structured verdict. The key selective-prediction behavior is that the agent may return \texttt{uncertain} rather than forcing a binary label when the evidence is not decisive.}
\label{fig:system_overview}
\end{figure*}

The experimental results in this paper are produced by the agent-based pipeline invoked by \texttt{run\_ece} in \texttt{eval/run\_eval.py}. Given a claim, the system instantiates a \texttt{VerificationAgent} implemented in \texttt{ece/agent.py}. Figure~\ref{fig:system_overview} summarizes the evaluated pipeline. The agent follows a tool-using workflow with up to eight interaction rounds:
\begin{enumerate}
    \item It first decides which tools to call for evidence gathering.
    \item It accumulates evidence from one or more external tools.
    \item It returns a structured \texttt{final\_answer} containing a verdict, confidence, source level, reasoning string, and optional source references.
\end{enumerate}

This is the system evaluated in the benchmark results below. The repository also contains lower-level heuristic modules for source classification and confidence storage, but those modules are not the primary pipeline used to generate the paper's benchmark numbers.

\subsection{Tool-Routed Evidence Gathering}

The agent uses four tools exposed through function calling: \nolinkurl{web_search} for general factual claims, \nolinkurl{arxiv_search} and \nolinkurl{semantic_scholar_search} for scholarly claims, and \nolinkurl{code_execute} for directly checkable computational claims.

The system prompt explicitly instructs the agent to start with web search, add scholarly search for research-related claims, and prefer executable verification for computational claims. This design is better characterized as \emph{tool routing plus LLM evidence synthesis} than as a fixed retrieval stack.

\subsection{Output Format and Abstention}

After gathering evidence, the agent emits a structured final answer with:
\begin{itemize}
    \item a ternary verdict (\texttt{confirmed}, \texttt{refuted}, or \texttt{uncertain});
    \item a confidence score in $[0,1]$;
    \item a source-level tag;
    \item a short reasoning field; and
    \item optional source references.
\end{itemize}

The \texttt{uncertain} verdict is the abstention mechanism studied in this paper. It is not produced by a separately implemented analytic uncertainty formula. Instead, the model directly decides whether the evidence is sufficient to commit to a binary verdict.

\subsection{Source-Level Semantics}

Source levels in the evaluated system follow the agent's implemented label set:
\begin{itemize}
    \item \textbf{L1}: code execution or other direct practical verification,
    \item \textbf{L2}: academic papers,
    \item \textbf{L3}: official documentation or books,
    \item \textbf{L4}: web pages, news, or forums,
    \item \textbf{L5}: model-internal knowledge without external evidence.
\end{itemize}

These source levels are assigned by the system during inference; they are \textbf{not} native annotations in the benchmark JSON. We therefore use them only for post-hoc analysis of the model's evidence profile and abstention behavior.

\section{Experiments}

\subsection{Benchmark and Baselines}

We evaluate on \textbf{ECE-Bench}, a benchmark of 95 claims across eight domains: science, technology, history, geography, medicine, controversial topics, mathematics, and misconceptions. The benchmark contains 57 true claims and 38 false claims. Its difficulty metadata uses three textual levels---\texttt{easy}, \texttt{medium}, and \texttt{hard}. It does \emph{not} contain a source-level field.

All four methods use the same model backend (GPT-5.4, \texttt{gpt-5.4-2026-03-05}). We compare:
\begin{itemize}
    \item \textbf{Vanilla}: direct LLM fact-checking without tools;
    \item \textbf{CoT}: chain-of-thought prompting without external retrieval;
    \item \textbf{Search}: LLM fact-checking with iterative web search;
    \item \textbf{ECE}: the tool-routed selective verification agent described above.
\end{itemize}

\subsection{Metrics}

We report standard accuracy, selective accuracy, coverage, macro-F1, ECE, Brier score, and AURC. For the three uncertainty metrics (ECE, Brier, AURC), lower is better. For the baselines, coverage is 100\%, so their selective accuracy is numerically identical to their standard accuracy.

\subsection{Main Results}

\begin{table}[t]
\centering
\scriptsize
\caption{Main results on ECE-Bench. ECE should be read as a selective-prediction system: it defers 6 of 95 claims, reaching 97.8\% accuracy on answered claims at 93.7\% coverage. Search remains the strongest baseline on standard accuracy and aggregate calibration metrics.}
\label{tab:main_results}
\setlength{\tabcolsep}{2pt}
\begin{tabular}{@{}lccccccc@{}}
\toprule
Method & Acc. & \shortstack{Sel.\\Acc.} & Cov. & \shortstack{Macro\\F1} & ECE$\downarrow$ & \shortstack{Brier\\$\downarrow$} & AURC$\downarrow$ \\
\midrule
Vanilla & 96.8\% & 96.8\% & 100.0\% & 0.968 & 0.0216 & 0.0310 & 0.0048 \\
CoT     & 96.8\% & 96.8\% & 100.0\% & 0.968 & 0.0149 & 0.0323 & 0.0096 \\
Search  & \textbf{97.9\%} & \textbf{97.9\%} & \textbf{100.0\%} & \textbf{0.978} & \textbf{0.0096} & \textbf{0.0213} & \textbf{0.0082} \\
ECE     & 91.6\% & 97.8\% & 93.7\% & 0.944 & 0.0508 & 0.0385 & 0.0182 \\
\bottomrule
\end{tabular}%
\end{table}

Table~\ref{tab:main_results} makes the trade-off explicit. Search is the strongest forced-answer baseline overall, achieving both the best standard accuracy and the best aggregate uncertainty metrics. ECE does \emph{not} outperform Search on those axes. Its contribution lies instead in selective behavior: it answers 89 of 95 claims and reaches 97.8\% accuracy on that answered subset.

This distinction is central to the paper's interpretation. Standard accuracy falls from the best baseline's 97.9\% to 91.6\% because abstentions are treated as non-correct outcomes. Selective accuracy, however, shows that once ECE elects to answer, its performance remains very strong. The appropriate conclusion is therefore one of selective abstention rather than global calibration superiority.

\subsection{Where ECE Defers}

ECE produces 6 \textit{uncertain} verdicts out of 95 claims. Five of these deferred cases are assigned source level L4 and one is assigned L2. The deferred set is evenly split between true and false claims (3 true, 3 false), so abstention should not be interpreted as a directional signal about the gold label. Instead, it indicates that the system found the available evidence insufficiently reliable or insufficiently decisive to justify a binary answer.

\begin{figure}[t]
\centering
\includegraphics[width=0.98\linewidth]{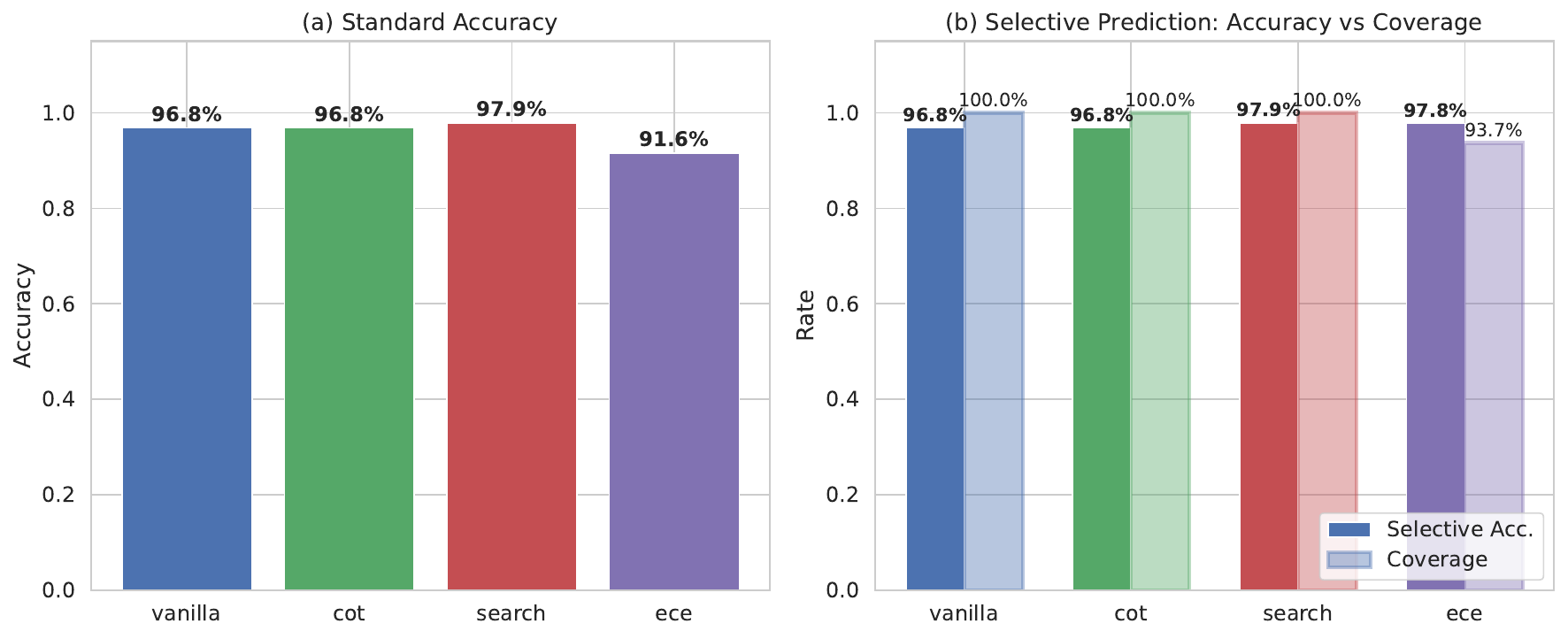}
\caption{Standard accuracy, selective accuracy, and coverage. ECE gives up a small amount of coverage to keep answered-claim accuracy high.}
\label{fig:accuracy_selective}
\end{figure}

Figure~\ref{fig:accuracy_selective} visualizes this trade-off directly. The baseline methods operate at full coverage, whereas ECE withholds 6 answers and nearly matches the strongest baseline's answered-claim accuracy.

\begin{figure}[t]
\centering
\includegraphics[width=0.95\linewidth]{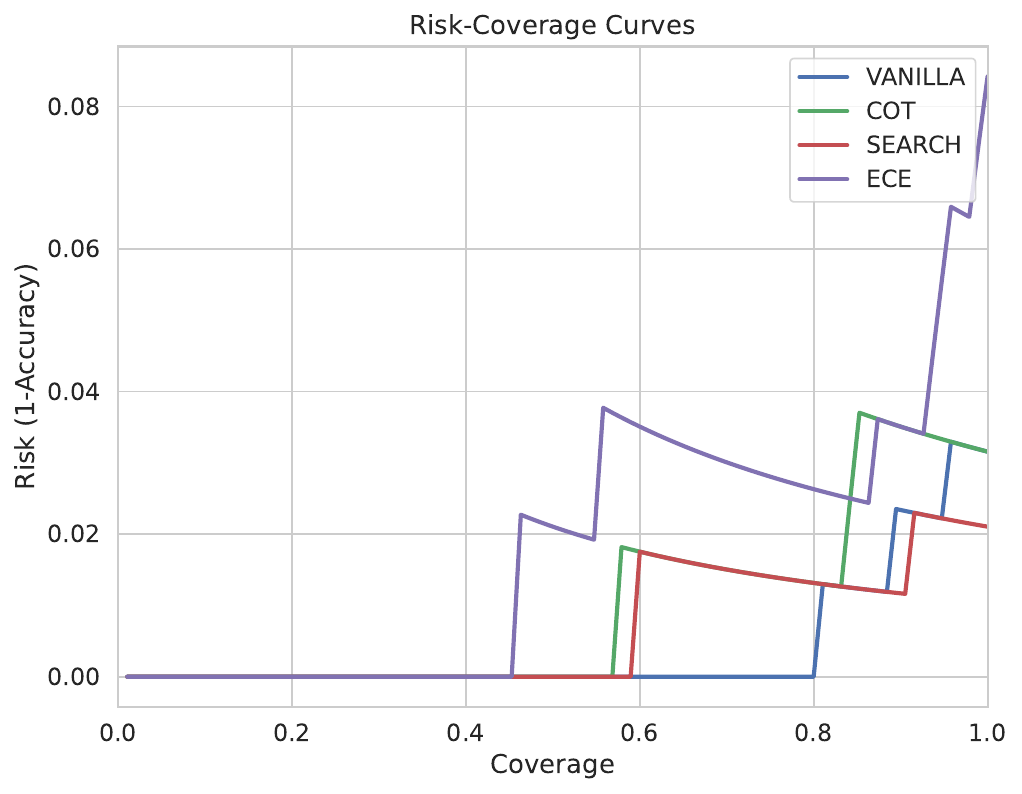}
\caption{Risk--coverage curves for all methods. The curve should be read jointly with Table~\ref{tab:main_results}: ECE is useful because it explicitly exposes a coverage--risk trade-off, not because it dominates all baselines on aggregate calibration metrics.}
\label{fig:risk_coverage}
\end{figure}

Figure~\ref{fig:risk_coverage} complements the table by showing how empirical risk evolves as predictions are ordered by confidence. This view is more faithful to the selective-prediction setting than any single top-line accuracy number.

\begin{figure}[t]
\centering
\includegraphics[width=0.98\linewidth]{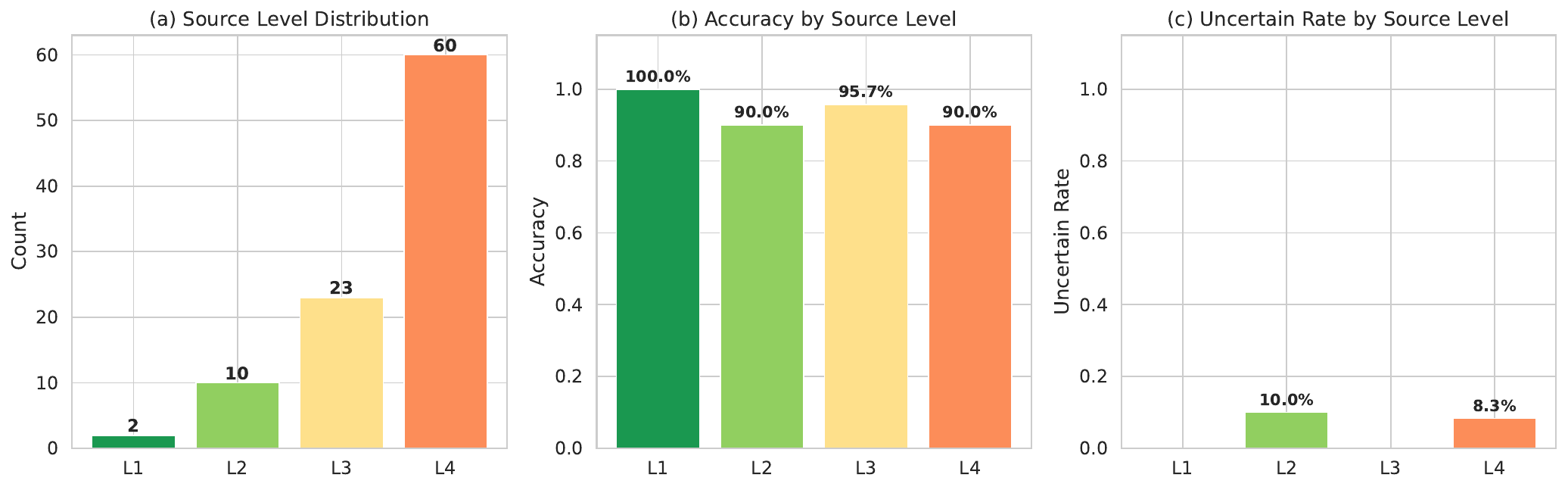}
\caption{ECE source-level analysis. Source levels are assigned by the system during inference rather than provided by the benchmark. Most deferred cases are concentrated in L4 evidence settings.}
\label{fig:source_levels}
\end{figure}

Figure~\ref{fig:source_levels} presents the post-hoc source-level analysis for ECE. Most claims are assigned to L4, and most abstentions also occur there. This pattern is consistent with a conservative strategy under weaker evidence conditions, although it also indicates that the current system relies heavily on web-level evidence.

\section{Discussion}

\subsection{Abstention as Safety-Oriented Behavior}

The main lesson from these experiments is not that ECE uniformly surpasses stronger retrieval baselines. Rather, ECE provides a transparent abstention mechanism for cases in which the evidence chain is too weak to support a definitive answer. In this sense, \textit{uncertain} is a safety-oriented output: it prevents low-quality evidence from being translated into a forced true/false claim.

This framing is especially important because the deferred cases are not overwhelmingly ``actually true'' or ``actually false.'' The correct empirical statement is simply that ECE deferred 6 claims, half of which have true gold labels. That result supports an epistemic-caution interpretation rather than any precision-style claim about the deferred subset.

\subsection{Limitations}

\begin{itemize}
    \item \textbf{Search remains stronger overall}: the Search baseline achieves higher standard accuracy and better ECE, Brier, and AURC values than ECE.
    \item \textbf{Source levels are system-generated}: source-level labels are assigned during inference rather than coming from benchmark annotation, so they should be treated as analysis metadata rather than ground truth.
    \item \textbf{Small benchmark}: ECE-Bench contains only 95 claims, which limits statistical power and domain coverage.
    \item \textbf{Implementation scope}: the evaluated system is a tool-routed agent, not a fully implemented dense retriever, reranker, or evidence-graph pipeline.
\end{itemize}

\subsection{Ethical Considerations}

By allowing abstention, ECE can reduce the risk of LLM-generated misinformation caused by overconfident false verdicts. At the same time, abstention can shift burden to downstream human reviewers, and system-generated source-level labels may themselves be imperfect. We therefore view ECE as a decision-support component rather than a fully autonomous fact-checker for high-stakes use.

\section{Conclusion}

We present Evidence Chain Evaluation (ECE) as a selective fact-checking system whose strength lies in deciding when not to answer. On ECE-Bench, ECE achieves 97.8\% selective accuracy at 93.7\% coverage by deferring 6 of 95 claims, with abstentions concentrated mainly in lower-reliability evidence settings. Although the strongest retrieval baseline remains better on standard accuracy and aggregate calibration metrics, ECE offers a practically meaningful trade-off between coverage and reliability. Future work should evaluate this abstention-oriented design on larger benchmarks and integrate stronger controls over evidence quality.

\bibliography{custom}

\clearpage
\appendix

\section{Additional Figures}
\label{sec:appendix}

\begin{figure}[H]
\centering
\includegraphics[width=0.95\linewidth]{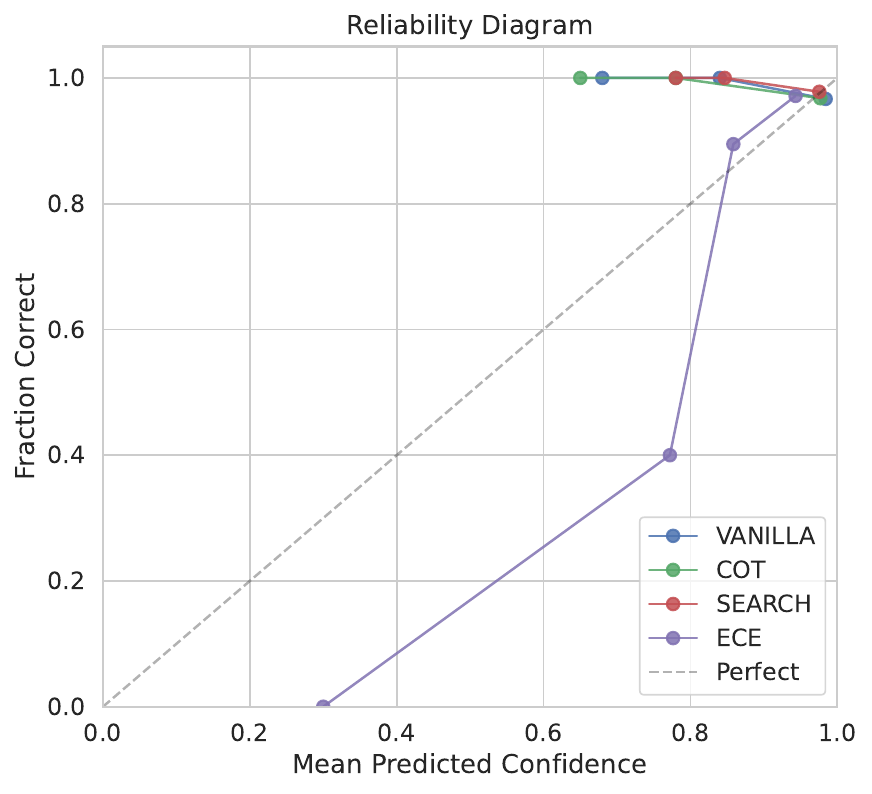}
\caption{Reliability diagrams for all methods. Search remains the strongest method on the aggregate calibration metrics reported in Table~\ref{tab:main_results}.}
\label{fig:calibration}
\end{figure}

\begin{figure}[H]
\centering
\includegraphics[width=0.98\linewidth]{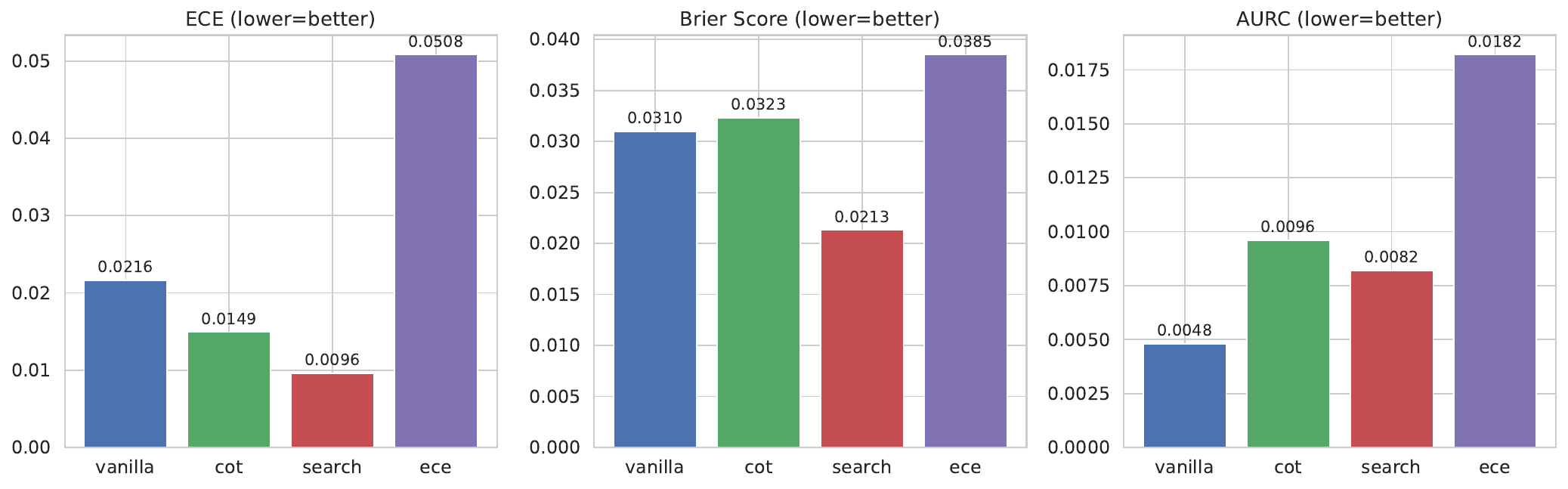}
\caption{Comparison of ECE, Brier score, and AURC. Lower is better for all three metrics.}
\label{fig:metrics}
\end{figure}

\begin{figure}[H]
\centering
\includegraphics[width=0.98\linewidth]{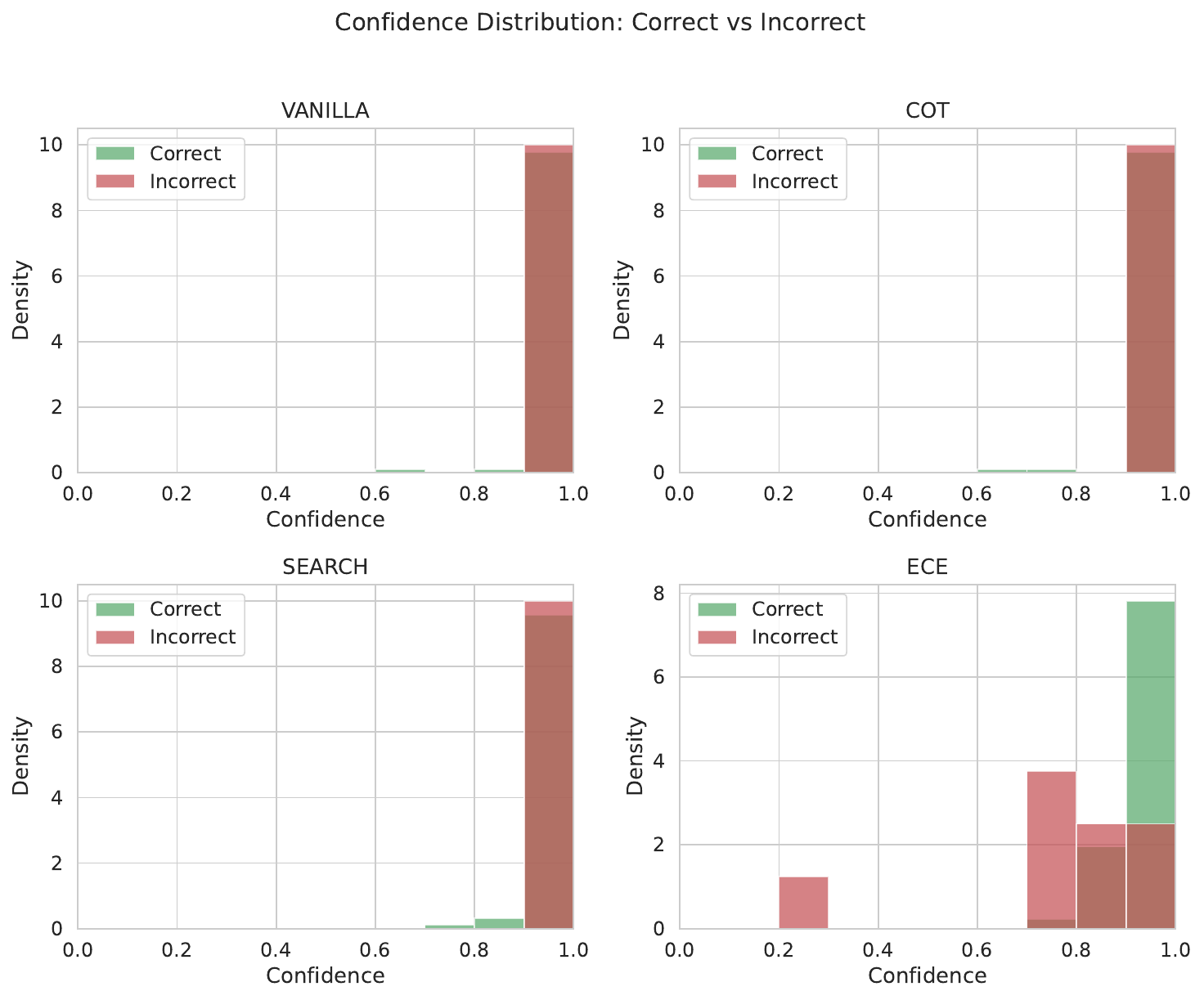}
\caption{Confidence distributions for correct and incorrect predictions across methods.}
\label{fig:confidence_dist}
\end{figure}

\end{document}